\documentclass[conference]{article}

\usepackage[margin=0.9in]{geometry}
\usepackage{algorithmic}
\usepackage{amsmath,amssymb,amsfonts}
\usepackage{booktabs}
\usepackage{cite}
\usepackage{graphicx}
\usepackage{makecell}
\usepackage{multirow}
\usepackage{placeins}
\usepackage{textcomp}
\usepackage{url}
\usepackage{xcolor}
\usepackage{xspace}
\usepackage{tikz}
\usetikzlibrary{positioning}

\usepackage[separate-uncertainty=true]{siunitx}
\begin{document}

% ----------------------------------------------------------------
\newcommand{\accuracy}{\ensuremath{a}\xspace}
\newcommand{\precision}{\ensuremath{p}\xspace}
\newcommand{\recall}{\ensuremath{r}\xspace}
\newcommand{\fone}{\ensuremath{F_1}\xspace}
\newcommand{\avg}[1]{\ensuremath{\overline{#1}}\xspace}
\newcommand{\starwars}{\textsc{starwars}\xspace}

\newcommand{\lang}{\ensuremath{\ell}\xspace}
\newcommand{\corpus}{\ensuremath{\mathcal{C}}\xspace}
\newcommand{\acorpus}{\ensuremath{\bar{\corpus}}\xspace}  % annotated corpus
\newcommand{\doc}{\ensuremath{d}\xspace}
\newcommand{\adoc}{\ensuremath{\bar{d}}\xspace}
\newcommand{\ndocs}{\ensuremath{n_d}\xspace}
\newcommand{\seq}[1]{\ensuremath{\langle #1 \rangle}\xspace}
\newcommand{\sent}{\ensuremath{s}\xspace}
\newcommand{\asent}{\ensuremath{\bar{\sent}}\xspace}
\newcommand{\word}{\ensuremath{w}\xspace}
\newcommand{\nsents}{\ensuremath{q}\xspace}
\newcommand{\nwords}{\ensuremath{r}\xspace}
\newcommand{\pair}[2]{\ensuremath{( #1, #2 )}\xspace}
\newcommand{\voc}{\ensuremath{V}\xspace}
\newcommand{\lab}{\ensuremath{t}\xspace}
\newcommand{\labels}{\ensuremath{T}\xspace}
\newcommand{\nlabs}{\ensuremath{n_t}\xspace}

% ----------------------------------------------------------------

\title{Novel Benchmark for NER in the Wastewater and Stormwater Domain}

\author{{Franco Alberto Cardillo}\\
\textit{CNR-ILC}\\
Pisa, Italy \\
francoalberto.cardillo@cnr.it
\and
{Franca Debole}\\
\textit{CNR-ISTI}\\
Pisa, Italy \\
franca.debole@cnr.it
\and
{Francesca Frontini}\\
\textit{CNR-ILC}\\
Pisa, Italy \\
francesca.frontini@cnr.it
\and
{Mitra Aelami}\\
\textit{HSM Univ. Montpellier}\\
\textit{IRD, CNRS, Inria}\\
Montpellier, France \\
mitra.aelami@umontpellier.fr
\and
{Nanée Chahinian}\\
\textit{ HSM Univ Montpellier} \\
\textit{IRD, CNRS}\\
Montpellier, France \\
nanee.chahinian@ird.fr
\and
{Serge Conrad}\\
\textit{ HSM Univ Montpellier} \\
\textit{IRD, CNRS}\\
Montpellier, France \\
serge.conrad@umontpellier.fr
}

% XXX Check date
\date{}

\maketitle

% ----------------------------------------------------------------

\begin{abstract}
Effective wastewater and stormwater management is essential for urban sustainability and environmental protection. Extracting structured knowledge from reports and regulations is challenging due to domain-specific terminology and multilingual contexts. This work focuses on domain-specific Named Entity Recognition (NER) as a first step towards effective relation and information extraction to support decision making. A multilingual benchmark is crucial for evaluating these methods. This study develops a French-Italian domain-specific text corpus for wastewater management. It evaluates state-of-the-art NER methods, including LLM-based approaches, to provide a reliable baseline for future strategies and explores automated annotation projection in view of an extension of the corpus to new languages.
\end{abstract}

% ----------------------------------------------------------------

% ----------------------------------------------------------------
\section{Introduction}\label{sec:introduction}
The effective management of wastewater and stormwater systems is crucial for urban sustainability and environmental protection. These systems, which form an integral part of public infrastructure, require structured information for monitoring, planning, and maintenance. However, much of the relevant information exists in unstructured textual formats, such as technical reports, regulatory documents, and maintenance logs. Extracting information from these sources is a key challenge, due to domain-specific terminology and the multilingual nature of regulatory and operational contexts.

Typically a wastewater management information extraction application will require domain-specific entity recognition, followed by the extraction of relations between entities to support decision-making, automated reasoning, and linking to existing knowledge bases. The recent progresses in domain-specific Named Entity Recognition (NER) have the potential to greatly facilitate the development of such applications. However, to effectively evaluate this first and crucial step of the extraction pipeline, it is essential to establish a clearly defined set of extractable entities and construct a \textbf{multilingual benchmark corpus}.

Building on previous work - carried out within the framework of a national project on just one language -  we propose the following contributions:
\begin{itemize}
    \item The \textbf{\starwars corpus}, an aligned French-Italian corpus containing domain-specific texts.
    \item Experimental evaluation of domain-specific NER method LLM-based approaches.
    \item Experimental evaluation of automated projection of annotations from one language to another, to facilitate the multilingual extension of the benchmark.
\end{itemize}

The resulting multilingual benchmark provides a foundation for further research and practical applications in the field of environmental information systems, first and foremost relation extraction (RE), and entity liking. However, such aspects will require further work. The remainder of this paper describes the corpus creation and annotation process, and NER experiments on both the Italian and French corpus, as well as the projection of annotation from one corpus to the other, and an analysis of the results.

% ----------------------------------------------------------------
\section{Related work}\label{sec:related-work}
Benchmarks, providing annotated corpora with domain-specific labels, as well as evaluation metrics, have been developed to test the effectiveness of NER in various domains. In the biomedical domain for instance, the BC5CDR corpus~\cite{10.1093/database/baw068} consists of 1500 PubMed articles with 4409 annotated chemicals, 5818 diseases and 3116 chemical-disease interactions. The SciERC~\cite{luan-etal-2018-multi} is a collection of 500 scientific abstracts annotated with scientific entities, their relations, and co-reference clusters. The BiodivNERE~\cite{10.3897/BDJ.10.e89481} is a gold-standard corpus specifically designed for NER and RE tasks within the biodiversity domain. This resource aims to enhance information extraction from biodiversity literature by providing annotated data tailored to the field's unique terminology and relationships. The corpus includes detailed annotations for entities such as organisms, phenomena, relying also on an underlying  ontology for classes and relations.

The emergence of such domain specific benchmarks is paralleled by the development of domain-tuned NER models, handling terminology and contexts in various fields. For example, the use of models such as BioBERT~\cite{btz682}, that are fine-tuned on domain corpora, can achieve high accuracy on field-specific entity types \cite{8965108}. More recently Large Language Models (LLMs) have begun to transform information extraction in technical and scientific domains, with the capacity to handle the various steps of entity extraction (extraction, normalisation, linking to existing knowledge) with little or no fine-tuning required. 

While some domains have been highly investigated, and with some evidence of good performances even with zero or few-shot approaches \cite{KOSPRDIC2024102970}, there is still a lack of exploration in water sciences. Some experiments on question answering, such as WaterGPT \cite{w16213075}, show that fine-tuning may be required. 

For the specific field of Wastewater networks, NER is further complicated by the lack of an agreed upon annotation schema and domain ontology, although the French Covadis standards\footnote{http://www.geoinformations.developpement-durable.gouv.fr/} constitutes an excellent starting point.  Multilinguality is another important dimension to consider, since texts containing relevant information are produced by local institutions in different languages. In general, available multilingual NER benchmarks such as MultiCoNER \cite{malmasi2022multiconerlargescalemultilingualdataset} do not contain texts and entities for all specific domains. Multilingual LLMs can be successfully applied in specialized fields (medicine, law, etc.) across different languages with or without fine-tuning, but performances may vary across language and yielding uneven results \cite{zhu2024multilinguallargelanguagemodels}.
%
% ----------------------------------------------------------------
\begin{table}
\begin{center}
\caption{The set of 14 tags in three languages}\label{tab:tagset}
\begin{tabular}{lll}
\toprule
 French & Italian & English \\
\midrule
Appareillage & Attrezzatura & Equipment \\
Spatial & Spaziale & Spatial \\
Evénéments & Evento & Event \\
Forme & Forma & Shape \\
Fonction & Funzione & Function \\
Defaillance & Guasto & Fault \\
Matériau & Materiale & Material \\
Mesure & Misura & Measurement \\
Mode\_circulation & Modalità\_circolazione & Circulation\_Mode \\
Organisation & Organizzazione & Organization \\
Ouvrage & Struttura & Structure \\
Temporel & Temporale & Temporal \\
Type\_reseau & Tipo\_rete & Network\_Type \\
% FIXME: substitute Pipeline with Conduit
Canalisation & Tubazione & Pipeline \\
\bottomrule
\end{tabular}
\end{center}
\end{table}
% ----------------------------------------------------------------

The most relevant datasets to date have been developed as part of the ``Megadata, Linked Data and Data Mining for Wastewater Networks'' (MeDo) project\footnote{\url{http://webmedo.msem.univ-montp2.fr/}}, funded by the Occitanie Region, in France, and carried out in collaboration between hydrologists, computational linguists and computer scientists. An information extraction platform was designed \cite{cherfi_weir-p_2021} and a corpus composed of 1,557 HTML and PDF documents was collected in July 2018 using a set of Google queries with a combination of keywords. A Gold standard containing the documents used to train the NER modules and the corresponding annotation files is available in an open access repository \cite{H0VXH0_2020}. While constituting a good basis, the corpus contains only documents in French and its spatial extent is limited to Metropolitan France.
% \FloatBarrier

% ----------------------------------------------------------------
\section{Corpus creation and annotation}\label{sec:corpus-creation}
The \textbf{\starwars Corpus}, currently in French and Italian, was created by consolidating and extending the MeDo corpus to obtain a multilingual parallel benchmark. 

% ----------------------------------------------------------------
\subsubsection{Annotation scheme}\label{sec:annotation-scheme}
The \starwars annotation scheme\footnote{Excerpts from the \starwars corpus alongside the annotation guidelines, describing the tags and their application in more detail, will be made available and cited in the final version of this paper.} is largely inspired by the UML schema of the RAEPA geostandard~\cite{geographique_geostandard_2019} and the application ontology developed by \cite{haydar_ontology_2024}. It contains 14 entities in total, 11 of which are common to RAEPA. 
The set of tags is in Table~\ref{tab:tagset}.

% ----------------------------------------------------------------
\begin{table}[t!]
\begin{center}
\caption{Corpus-level Statistics}\label{tab:corpus-level-stats}
\begin{tabular}{l%
S[table-format=5.0, input-ignore={,}] %
S[table-format=5.0, input-ignore={,}]}
\toprule
 & \multicolumn{1}{c}{French} & \multicolumn{1}{c}{Italian} \\
 \midrule
 Documents & 110 & 110 \\
 Sentences & 3,103 & 3,103 \\
Annotated sentences (with 1+ tags) & 1726 & 1726 \\
 Tokens &77,530 & 72,972 \\
 Annotated tokens (either B- or I-) & 12,415 & 13,470 \\
 Entities (tokens annotated with B-) & 5,357 & 5,324 \\
 \bottomrule
\end{tabular}
\end{center}
\end{table}
% ----------------------------------------------------------------

% \FloatBarrier

% ----------------------------------------------------------------
\subsubsection{French Corpus}\label{sec:french-corpus}
The French corpus consists of 110 documents, 45 of which are from the MeDo Gold Standard, while the remaining were collected successively. It mostly consists of newspaper articles and official documents describing works on the network. The covered areas are the municipality of Montpellier, N\^imes and Al\`es. All the annotations were manually checked and updated to align with the new annotation scheme.

% ----------------------------------------------------------------
\subsubsection{Italian Corpus}\label{sec:italian-corpus}
The Italian corpus was obtained by machine translating the French texts; two translation students (Italian native speakers with excellent command of French) checked the translations independently; in case of ambiguity, expert hydrologists identified the correct Italian translation. 

% ----------------------------------------------------------------
\subsubsection{Annotation alignment and format}
Once the Italian corpus was manually checked, both corpora were uploaded to an instance of the INCEpTION tool\footnote{\url{https://inception-project.github.io/}} and the second student manually transposed the annotation from the French corpus onto the Italian one.  %INCEpTION allows for export in various annotation formats, including the widely used IOB \cite{ramshaw-marcus-1995-text}.

% ----------------------------------------------------------------
\subsubsection{Corpus statistics }
Statistical characteristics for each language: in Table~\ref{tab:corpus-level-stats}
some global statistics at corpus-level, in Table~\ref{tab:doc-level-stats} statistics at document level, the minimum $m$, the maximum $M$, the mean $\mu$ and the standard deviation $\sigma$ 
%of the number of tokens, the number of sentences, the number of annotated tokens, the number of unique tags and the number of tags per document, in Table~\ref{tab:tag-level-stats} the statistics at tag-level.
of the number of tokens / sentences / annotated tokens /tags per document/unique tags, in Table~\ref{tab:tag-level-stats} the statistics at tag-level.

% ----------------------------------------------------------------
% \input{tex_files/table-doc-level-stats}
\begin{table}[t!]
\caption{Document-level Statistics. $m$, $M$, $\mu$, $\sigma$ represent, respectively, the min, max, mean and standard deviation.}\label{tab:doc-level-stats}
\begin{center}
\begin{tabular}{@{}p{1.5cm} %
    S[table-format=3] @{\hspace{0.05em} --\hspace{0.05em}}@{} @{}S[table-format=4]
    S[table-format=3.2] @{\hspace{0.05em}\tiny{$\pm$}\hspace{0.05em}} S[table-format=3.2]@{}
    S[table-format=3] @{\hspace{0.05em}--\hspace{0.05em}} S[table-format=4] 
    S[table-format=3.2] @{\hspace{0.05em}\tiny{$\pm$}\hspace{0.05em}} S[table-format=3.2]}
\toprule
    & \multicolumn{4}{c}{French} & \multicolumn{4}{c}{Italian} \\
\cmidrule(lr){2-5}
\cmidrule(lr){6-9}
 & \multicolumn{2}{c}{$m$ -- $M$} & \multicolumn{2}{c}{$\mu \pm \sigma$} & \multicolumn{2}{c}{$m$ -- $M$} & \multicolumn{2}{c}{$\mu \pm \sigma$} \\
\midrule
Sentences & 2 & 136 & 28.21 & 24.99 & 2 & 136 & 28.21 & 24.99 \\
Tokens & 54 & 3941 & 704.82 & 598.57 & 46 & 3726 & 663.38 & 563.20 \\
$\hookrightarrow$ \scriptsize{Annot.}  & 5 & 901 & 112.86 & 114.73 & 7 & 838 & 122.45 & 116.38 \\
Tags & 3 & 446 & 48.70 & 53.33 & 3 & 431 & 48.40 & 52.27 \\
$\hookrightarrow$\scriptsize{Unique} & 1 & 12 & 7.63 & 2.11 & 1 & 12 & 7.63 & 2.11 \\
\bottomrule
\end{tabular}

\end{center}
\end{table}
% ----------------------------------------------------------------

\section{Experiments}\label{sec:experiments}
The objective of the experiments presented in this section is twofold: first, to establish a baseline for each specific experimental task, which will serve as a reference for the design and evaluation of improved and more advanced NER methods
applied to the \starwars corpus; and second, to assess the manual effort and computational resources required to produce high-quality annotated corpora under low-resource settings, such as the \starwars domain. We conduct three types of experiments.
The first investigates the manual effort required to build an automatic annotator for a monolingual corpus. Specifically, we investigate how many manually annotated monolingual documents are needed to fine-tune models in the BERT family, pre-trained in either French or Italian, for automatically annotating the \starwars corpus.
The second examines the feasibility of transferring annotations from a monolingual corpus to sentence-aligned translations of its documents. We formulate this task as annotation projection between corresponding tokens and text spans.
The third explores the use of recent very large language models in a zero-shot setting for annotation projection, using the same formulation of the previous experiment. 

% ----------------------------------------------------------------
%\input{tex_files/table-tag-level-stats}
\begin{table}[t!]
\begin{center}
\caption{Tag-level Statistics}\label{tab:tag-level-stats}
\begin{tabular}{l
S[table-format=3.0, input-ignore={,}]
S[table-format=3.0, input-ignore={,}]
S[table-format=3.0, input-ignore={,}]
S[table-format=3.0, input-ignore={,}]
S[table-format=3.0, input-ignore={,}]
S[table-format=3.0, input-ignore={,}]
} %
\toprule
 & \multicolumn{3}{c}{French} & \multicolumn{3}{c}{Italian} \\
 \cmidrule(lr){2-4}
 \cmidrule(lr){5-7}
Tag & \multicolumn{1}{c}{Docs} & \multicolumn{1}{c}{Ent} & \multicolumn{1}{c}{Tokens} & 
        \multicolumn{1}{c}{Docs} & \multicolumn{1}{c}{Ent} & \multicolumn{1}{c}{Tokens} \\
\midrule
Circulation\_Mode & 20 & 44 & 56 & 20 & 43 & 72 \\
Equipment & 23 & 40 & 87 & 23 & 39 & 97 \\
Event & 87 & 519 & 768 & 87 & 520 & 862 \\
Fault & 33 & 171 & 438 & 33 & 164 & 409\\
Function & 39 & 83 & 86 & 39 & 80 & 131\\
Material & 29 & 102 & 168 & 29 & 103 & 229 \\
Measurement & 67 & 386 & 1138 & 67 & 383 & 1179 \\
Network\_Type & 96 & 712 & 1659 & 96 & 707 & 1822 \\
Organization & 82 & 379 & 1199 & 82 & 379 & 1238  \\
Pipeline & 81 & 367 & 487 & 81 & 362 & 640 \\
Shape & 7 & 8 & 11 &  7 & 8 & 11\\
Spatial & 99 & 1409 & 3401 & 99 & 1397 & 3571 \\
Structure & 89 & 669 & 1632 & 89 & 667 & 1742 \\
Temporal & 87 & 468 & 1282 & 87 & 472 & 1464\\
\bottomrule
\end{tabular}
\end{center}
\end{table}
% ----------------------------------------------------------------
%
We denote the French and the Italian corpus with, respectively, $\corpus^{\textsc{F}}$ and $\corpus^{\textsc{I}}$. Each $\corpus^{\ell}$ consists of a set of \ndocs documents $\left\{ \doc^{\lang}_{i} \mid i=1..\ndocs\right\}$, where each document $\doc^{\ell}_{i} =\seq{ \sent^{\lang}_{i,j} \mid j=1..\nsents^{l}_{i,j}}$ is a sequence of sentences. A sentence $\sent^{l}_{i,j} =
        \seq{
                \word^{\lang}_{i,j,k} \mid k=1..\nwords^{l}_{i,j}
            }$
consists of a sequence of tokens\footnote{Tokens include words, numbers, punctuation, etc.} $w^\lang$ from the vocabulary $\voc^\lang$ of the language $\lang$. We will use $\acorpus^\lang$,
$\adoc^\lang$ and $\asent^\lang$ for denoting, respectively, an annotated corpus, document and sentence in the language \lang. Annotated sentences are sequences
$\asent^{\lang}_{i,j} =
        \seq{
                \pair{ \word^{\lang}_{i,j,k} }{ \lab^{\lang}_{i,j,k} } \mid k=1..\nwords^{\lang}_{i,j}
            }$
consisting of pairs whose elements are a token $\word^\lang$ and an associated tag $\lab^\lang$. Hereafter, we omit indices whenever there is no ambiguity. The sentences $\asent^\lang$ are annotated according to the IOB2 annotation scheme~\cite{TjongKimSangBuchholz2000Introduction}, requiring the first token of a multi-token annotation of type T to be annotated as \texttt{B-T}. 
In these experiments, in order to better simulate a realistic use case, texts in both $\corpus^{\textsc{F}}$ and $\corpus^{\textsc{I}}$ are neither normalized nor cleaned, meaning that no noise removal or correction of spelling and grammatical errors is performed. Moreover, we do not control the prevalence of labelled tokens and sentences.

% ----------------------------------------------------------------
\subsection{Automatic annotation}\label{sec:automatic-annotation}
Since the manual annotation of technical domains requires significant domain expertise and time, we investigate whether and to what extent the fine-tuning of pre-trained LLMs can substitute a human expert in the document annotation task.

Specifically, we frame this task as the induction of an annotation function that assigns one of the tags in Table~\ref{tab:tagset} to input tokens. As a baseline approach, we select variants of the BERT model~\cite{devlin2019bert}, an encoder-only architecture commonly used for token-level annotations tasks, each pre-trained either in French or Italian. To this end, we partition the annotated corpus $\acorpus^\lang$ into training, validation and test sets and fine-tune the language-specific pre-trained model in a token classification task according to the IOB2 scheme, using the expert annotations in $\acorpus^\lang$ as target classes in this supervised learning task. 

For $\acorpus^{\textsc{F}}$, we use CamemBERT~\cite{Martin2020} (110M parameters, pre-trained on 138GB with 32.7B tokens), which is a well-established model for French texts.
Since Italian lacks an equivalent model, for $\acorpus^{\textsc{I}}$ we test two published models: Italian BERT\footnote{\url{https://github.com/dbmdz/berts}} (110M, 13GB/2B tokens); GilBERTo\footnote{\url{https://github.com/idb-ita/GilBERTo}} (110M,71GB/11B tokens).

We conduct the experiments using a single 70\%--10\%--20\% split of the $\acorpus^\lang$ into training/validation/test sets. The three partitions are aligned across the languages, meaning that the French and Italian sets contain the same documents. Since we aim at evaluating the model performance as a function of the amount of annotated data, we train the models on a strictly increasing sequence of training subsets. For each training size we perform multiple runs. The test remains fixed across all the training runs.

% ---------------------------------------------------------------------------
\subsection{Annotation projection}\label{sec:annotation-projection}
Given a pair of parallel corpora consisting of an annotated source corpus $\acorpus^{s}$ and an unannotated target corpus $\corpus^{t}$, 
the task of annotation projection can be defined as transporting the labels 
from the annotated text spans in $\acorpus^s$ to the corresponding text 
spans in $\corpus^t$~\cite{garcia2022t}. We conduct experiments using either $\acorpus^F$ or $\acorpus^I$ as the source corpus and projecting its annotations 
to the unannotated target corpus $\corpus^I$ or $\corpus^F$, respectively. 
We then evaluate the quality of the projection comparing the final tagging of a target sentence $\sent^t$ against the golden annotations in $\asent^{s}$.
As the number and order of words may differ across aligned sentences $\asent^s$ and $\sent^t$, we use two LLM-based methods for the automatic computation of word alignments: SimAlign~\cite{sabet2020simalign} and AWESoME-align~\cite{DouNeubig2021Worda} (hereafter, AWESoME-a), both of which have been shown to outperform classical statistical-based alignment methods%
\footnote{\url{https://en.wikipedia.org/wiki/IBM_alignment_models}}. 
SimAlign computes the word alignments by applying several heuristics to the similarity degree of either static or contextualized embeddings computed at either the word or sub-word level. SimAlign uses fastText~\cite{bojanowski2017enriching} for the static embeddings and either mBERT~\cite{devlin2019bert} or XLM-R~\cite{conneau2020unsupervised}, pre-trained, 
respectively, on 104 and 100 languages -- including French and Italian, for contextualized embeddings.
AWESoME-a improves upon SimAlign by fine-tuning the pre-trained LLM 
on down-stream tasks specifically designed to increase the similarity of 
contextual embeddings from the source and target sentences. AWESoME-a does not use any heuristics in the alignment process. 
In our experiments, both SimAlign and AWESoME-a use contextual embeddings at the sub-word level computed by mBERT. Additionally, we configure SimAlign 
to use the three heuristics \textsf{mwmf}, \textsf{inter}, \textsf{itermax} -- see~\cite{sabet2020simalign} for details. Given a pair of source and target sentences and an alignment between their terms, we annotate each target term occurring in the alignment with the tag of the corresponding source term. The target terms not in the alignment are annotated with \texttt{"O"}.

% ----------------------------------------------------------------
% \input{tex_files/table-frenchresults}
\begin{table}[t!]
\begin{center}
\caption{French Automatic Annotation}\label{tab:frenchresults}
    \begin{tabular}{
    @{}S[table-format=1.2] 
    S[table-format=1.2] @{\hspace{0.05em}$\pm$\hspace{0.05em}} S[table-format=1.2]
    S[table-format=1.2] @{\hspace{0.05em}$\pm$\hspace{0.05em}} S[table-format=1.2]
    S[table-format=1.2] @{\hspace{0.05em}$\pm$\hspace{0.05em}} S[table-format=1.2]
    S[table-format=1.2] @{\hspace{0.05em}$\pm$\hspace{0.05em}} S[table-format=1.2]
    S[table-format=1.2] @{\hspace{0.05em}$\pm$\hspace{0.05em}} S[table-format=1.2]
    S[table-format=1.2] @{\hspace{0.05em}$\pm$\hspace{0.05em}} S[table-format=1.2]}
    \hline
        \toprule   & \multicolumn{6}{c}{CamemBERT}\\
        \cmidrule(lr){2-7}
        \cmidrule(lr){8-13}
        Ratio  & \multicolumn{2}{c}{$\avg{\precision} \pm \sigma$}  &  \multicolumn{2}{c}{\avg{\recall}$\pm\ \sigma$}  &  
        \multicolumn{2}{c}{$\avg{\fone} \pm\ \sigma$} \\ 
        \midrule
        0.10 & 0.31 & 0.06 & 0.42 & 0.10 & 0.35 & 0.07  \\ 
        0.25 & 0.46 & 0.01 & 0.59 & 0.02 & 0.52 & 0.01  \\
        0.50 & 0.55 & 0.01 & 0.65 & 0.01 & 0.59 & 0.01  \\ 
         1.00 & 0.57 & 0.02 & 0.68 & 0.02 & \textbf{.63} & \textbf{.02}  \\ 
        \bottomrule
     \end{tabular}
\end{center}
\end{table}
%
% ----------------------------------------------------------------
% \input{tex_files/table-italianresults}
\begin{table}[t!]
\caption{Italian Automatic Annotation}\label{tab:italianresults}
\centering
{
    \begin{tabular}{@{}S[table-format=.2] 
    S[table-format=.2] @{\hspace{0.05em}$\pm$\hspace{0.05em}} S[table-format=.2]
    S[table-format=.2] @{\hspace{0.05em}$\pm$\hspace{0.05em}} S[table-format=.2]
    S[table-format=.2] @{\hspace{0.05em}$\pm$\hspace{0.05em}} S[table-format=.2]
    S[table-format=.2] @{\hspace{0.05em}$\pm$\hspace{0.05em}} S[table-format=.2]
    S[table-format=.2] @{\hspace{0.05em}$\pm$\hspace{0.05em}} S[table-format=.2]
    S[table-format=.2] @{\hspace{0.05em}$\pm$\hspace{0.05em}} S[table-format=.2]}
    \hline
        \toprule
        & \multicolumn{6}{c}{ItalianBert} & \multicolumn{6}{c}{GilBERTo} \\
        \cmidrule(ll){2-7}
        \cmidrule(ll){8-13}
        Ratio  & \multicolumn{2}{c}{$\avg{\precision} \pm \sigma$}  &  \multicolumn{2}{c}{$\avg{\recall}\pm\ \sigma$}  &  
        \multicolumn{2}{c}{$\avg{\fone} \pm\ \sigma$} 
        & \multicolumn{2}{c}{$\avg{\precision} \pm \sigma$}  &  \multicolumn{2}{c}{$\avg{\recall}\pm\ \sigma$}  &  
        \multicolumn{2}{c}{$\avg{\fone} \pm\ \sigma$} 
        \\ 
        \midrule
        .10 & .29 & .01 & .29 & .06 & .28 & .03 & .27 & .03 & .28 & .02 & .27 & .02\\ 
        .25 & .43 & .02 & .42 & .03 & .43 & .03 & .43 & .02 & .40 & .06 & .41 & .02\\
        .50 & .50 & .04 & .49 & .02 & .49 & .01 & .49 & .03 & .47 & .01 & .48 & .02 \\
         1.00  & .53 & .01 & .52 & .02 & \textbf{.53} & \textbf{.01} & .51	& .04 & .50	& .04 & \textbf{.50}	& \textbf{.01}\\
        \bottomrule
     \end{tabular}
}
\end{table}
%
% ----------------------------------------------------------------
% \FloatBarrier
\subsection{Annotation Projection by LLMs}\label{sec:generative-llm}
We also include a preliminary investigation into the use of LLMs for annotation projection. We experiment with open-weight, multilingual, decoder-only LLM families, namely Llama2~\cite{touvron2023llama}, LLama3~\cite{grattafiori2024llama}, Mistral\footnote{\url{https://mistral.ai}} and Phi4~\cite{abdin2024phi}. We consider models ranging in size from 
1.2 to 122.6 billion parameters\footnote{We use quantized versions of the largest ones distributed via the Ollama tool -- \url{https://ollama.com/}.}, listed in Table~\ref{tab:results-annotation-projection-llm}. 
We design two experimental tasks that differ in the annotation scheme. In the first task, the sentences are annotated using the IOB2 format, as in the previous experiments.
In the second task, following~\cite{garcia2022t}, we annotate the sentences with HTML-like tags. In fact, since the \starwars annotation guidelines do not allow the nesting of tags, an IOB2 annotation like ``$w_i$ \texttt{B-T} $w_{i+1}$ \texttt{I-T}'' can also be expressed as $\langle T \rangle w_i w_{i+1} \langle /T \rangle$. We use the same prompt for both tasks, 
%``asking'' the LLMs 
requesting to project the annotations from the annotated source to the unannotated target, without introducing new entities or including comments or reasoning steps. Each task is run twice, one for each source/target language pair.

%\vspace{0.5em}
We acknowledge the need for a more extensive evaluation in the three experiments, but we had to scale them according to the available computational resources and budget.

% ----------------------------------------------------------------
\section{Evaluation and Discussion}\label{sec:evaluation}
NER can be evaluated with two different approaches: token-level and entity-level. In the latter approach, the evaluation treats all tokens within a multi-token entity as a single unit, while the former, the token's classification is evaluated independently. Since in NER the focus of the evaluation is on how well the model predicts entire entities rather than individual tokens, we use the entity-level approach, following the standard practice in sequence labelling evaluation~\cite{TjongKimSangBuchholz2000Introduction}. For this reason, we compute the results in this section using the Python package \texttt{seqeval}\cite{seqeval}, configured in the ``default'' evaluation mode, which corresponds to the original CoNLL\footnote{https://www.conll.org/} evaluation script. We evaluate and report all  the metrics (precision \precision, recall \recall and \fone scores), using micro-averaging and select the \fone score as the primary metric for model selection.
\subsection{Results on Automatic Annotation}
We fine-tune the three models with (increasing) 
subsets corresponding to 10\%, 25\%, 50\% and 100\% of the full training
set. For each subset, we repeat the fine-tuning three times using different
random seeds in model initialization and data shuffling. For each subset, 
the models are fine-tuned for 40 epochs, with their performance 
evaluated on the validation set at the end of each epoch. 
Once the training ends, we use the parameters with the
best validation performance for evaluating the model on the test set. We
remark here that the all the sets (and subsets) are
aligned and contain the same documents in both languages.
%
% ----------------------------------------------------------------
% \input{tex_files/table-anno-projection}
\begin{table}[t!]
\caption{Results of Annotation Projection}\label{tab:results-annotation-projection}
\newcommand{\aplastcol}{8}
\begin{center}
    \begin{tabular}{llcccccc}
        \toprule
         & & \multicolumn{3}{c}{French $\rightarrow$ Italian} & \multicolumn{3}{c}{Italian $\rightarrow$ French} \\
         \cmidrule(lr){3-5} \cmidrule(lr){6-\aplastcol} 
        Method &   & \precision & \recall & \fone & \precision & \recall & \fone \\
        \midrule
        \multirow{3}{*}{SimAlign} & \textsf{inter}  & .62 & .65 & .63  & .60 & .66 & .63  \\
         & \textsf{itermax}   & .62 & .66 & .64  & .58 & .66 & .62 \\ 
         & \textsf{mwmf}   & .59 & .63 & .61 &  .54 & .63 & .58 \\
        \cmidrule{2-\aplastcol} 
        \multicolumn{2}{l}{AWESoME-a}  & .63 & .65 & .64 &  .60 & .66 & .63 \\
        \cmidrule{2-\aplastcol}
        \multirow{4}{*}{\makecell{AWESoME-a}}  
            & epochs:\hfill\ 1 & .62 & .65 & .63 & .59 & .66 & .62 \\
            & \hfill\ 2 & .63 & .66 & .64 & .60 & .67 & .63 \\
             & \hfill\ 3 & .63 & .66 & .64 & .60 & .67 & .63 \\
            with & \hfill\ 5 & .64 & .67 & .65 & .60 & .67 & .64 \\
            fine-tuning & \hfill\ 10 & .64 & .67 & \textbf{.66} & .61 & .68 & .64 \\
            & \hfill\ 25 & .64 & .68 & .66 & .61 & .69 & \textbf{.65} \\
            & \hfill\ 50 & .64 & .68 &.66 & .62 & .69 & .65 \\
        \bottomrule
    \end{tabular}
\end{center}
\end{table}
%
% ----------------------------------------------------------------
% \FloatBarrier 
Table~\ref{tab:frenchresults} shows the results for CamemBERT on $\acorpus^{\textsc{F}}$ and includes the average ($\avg{\precision}$, $\avg{\recall}$, $\avg{\fone}$ ) and standard deviation ($\sigma$) of each metric over the three
runs. Table~\ref{tab:italianresults} contain the same values for the two BERT models
on $\acorpus^{\textsc{I}}$. First, we note that this experiment is not conclusive. Since all the models reach their maximum average \fone, $0.63$, $0.53$ and $0.50$ respectively for CamemBERT, ItalianBERT and GilBERTo, using 100\% of the training data, their generalization error would likely decrease with larger training sets. We can thus conclude that we need to collect more annotated data for both languages. Second, there is a huge gap in the model performance between $\acorpus^{\textsc{F}}$ and $\acorpus^{\textsc{I}}$ ($0.63$ vs $0.53$). Some errors made by the Italian models can be attributed to the translation process, which occasionally left untranslated terms (e.g., \textit{piquages}) or made multiple translations of the same terms (e.g., \textit{boîte de branchement} translated as \textit{camera di allaccio/scatole di derivazione}).

However, this performance gap can be attributed also to the much larger pre-training dataset used for CamemBERT (32.7B vs 2B and 11B tokens for the Italian models), which likely results in more expressive contextual embeddings, better suited to capture the nuances of technical documents, and which potentially includes also more domain-specific terms than the Italian ones. A deeper analysis of the Italian results using the metrics defined in the SemEval-2013 task 9~\cite{semeval13}, not included here due to space constraints, shows that the number of \textit{missed} is higher than the number of \textit{incorrect} ones ($25\%$ vs $20\%$)%
\footnote{\textit{Missed}: annotated in the golden but not by the system; \textit{Incorrect}: wrong annotation.}.%
Additionally, \textit{missed} annotations more frequently involve low-frequency tags. These findings are consistent with the interpretation that limited pre-training data lead to poorer representations and reduced performance on domain-specific texts and they are corroborated by the Italian results in annotation projection discussed in the next section. All the models are fine-tuned using the Adam optimizer, a $10^{-5}$ learning rate, 16 batch size, leaving the remaining hyperparameters to the recommended value.

% ----------------------------------------------------------------
\subsection{Annotation projection}\label{sec:annotation-projection-res}
In all experiments, we use mBERT as the base model for computing the contextual embeddings used by SimAlign and AweSome-a. Concerning the latter, we report results both without and with fine-tuning on a range of epochs while keeping the other hyperparameters at their recommended values. The results in Table~\ref{tab:results-annotation-projection} show that, without fine-tuning, the two methods have comparable performance, although, in a real-world scenario, it would be unclear how to select the optimal heuristics for SimAlign. After fine-tuning, AweSome-a slightly outperforms SimAlign, confirming the results in~\cite{DouNeubig2021Worda}. For both language pairs, the fine-tuned AweSome-a outperforms the mono-lingual models: $0.66$ \fone vs $0.63$ for French, and $0.65$ vs $0.50$ for Italian. For Italian, where the tested pre-trained BERT models struggle with the technical jargon of this corpus, using them for text annotation is ineffective, whereas projecting annotations from French yields an \fone comparable to that of the monolingual model on the French subset. These results further support the conclusion that the two Italian BERT models used in the previous experiment build very poor contextual embeddings. In fact, the contextual embeddings for Italian built by mBERT enable an alignment with French with no performance gap between the two alignment directions.

% ----------------------------------------------------------------
%\input{tex_files/table-llm-anno-projection}
\begin{table}[t!]
\newcommand{\tHTML}{\footnotesize{\textsf{HTML}}}
\newcommand{\tIOB}{\footnotesize{\textsf{IOB2}}}
\newcommand{\aplastcol}{11}
\renewcommand\cellgape{}
\newcommand{\tightstack}[1]{\makecell[{{c}}]{#1}}

\caption{Annotation Projection by LLMs}\label{tab:results-annotation-projection-llm}
\begin{center}
    \begin{tabular}{l@{\hspace{0.1em}}r@{\hspace{0.5em}}ccccccccc}
        \toprule
         & & & \multicolumn{4}{c}{French $\rightarrow$ Italian} & \multicolumn{4}{c}{Italian $\rightarrow$ French} \\
         \cmidrule(lr){4-7} \cmidrule(lr){8-\aplastcol} 
        LLM & Size & Task & Tags & \precision & \recall & \fone & Tags & \precision & \recall & \fone \\
        \midrule
        Gemma3
        & 27.4  & \tHTML & \textbf{14} & .68 & .50 & \textbf{.57}  & \textbf{14} & .74 & .48 & \textbf{.58}\\  
         &  & \tIOB & 15 & .69 & .56 &.62  & \multicolumn{4}{c}{Runs always fail}  \\
        \cmidrule{3-\aplastcol}
        Llama2 & 69.0 
         & \tHTML & 41 & .62 & .30 &.41  & 26 & .70 & .29 & .41  \\  
         &  & \tIOB & 21 & .59 & .26 &.36  & 15 & .60 & .13 & .22  \\
        \cmidrule{3-\aplastcol}
        Llama3 & 1.2 & \tHTML &  \multicolumn{8}{c}{\scriptsize{Unpredictable responses, syntactical issues}} \\
         &  & \tIOB & 110 & .21 & .04 & .07 & 119 & .19 & .03 & .06 \\
        \cmidrule{3-3}
          & 3.2 & \tHTML & 57 & .47 & .18 &.26 
        & 45 & .52 & .16 & .25\\
         &  & \tIOB & 121 & .41 & .21 & .28 & 108 & .35 & .12 & .18  \\
         \cmidrule{3-3}
         %
        % llama architecture
         & 70.6
        & \tHTML & 38 & .66 & .51 & .58  & 24 & .77 & .55 & .64  \\  
         &  & \tIOB & \textbf{14} & .70 & .64 & \textbf{.67} & 15 & .73 & .53 & \textbf{.62}  \\
         \cmidrule{3-\aplastcol}
        Mistral & 7.25
        % mistral
         & \tHTML & 60 & .61 & .27 & .37 &  42 & .65 & .19 & .29 \\
         &  & \tIOB & 45 & .58 & .17 & .26 & 47 & .54 & .11 & .19  \\
        \cmidrule{3-3}
         & 23.6
         % mistral-small
         & \tHTML & 21 & .68 & .46 & .55  & 25 & .74 &.47 & .58  \\ 
         &  & \tIOB & 15 & .68 & .52 & .59 &  15 & .68 & .27 & .39 \\
         \cmidrule{3-3}
          & 122.6 & \tHTML & \textbf{14} & .70 & .40 & .51 & \textbf{14} & .74 & .41 & .53  \\ 
         &  & \tIOB & 16 & .70 & .61 & .64 & 15 & .70 & .48 & .57  \\
         \cmidrule{3-\aplastcol}
        Phi4 & 14.7
         & \tHTML & 16 & .66 & .45 & .53  & 16 & .71 &.41 & .52  \\ 
         &  & \tIOB & 26 & .67 & .51 & .58 & 27 & .67 & .30 & .41  \\
        \bottomrule
    \end{tabular}
\end{center}
\end{table}
%

% ----------------------------------------------------------------
\subsection{Annotation Projection by LLMs}\label{sec:annotation-projection-llm}
We experimented with several models, in different families and varying sizes, and include the results for nine of them in Table~\ref{tab:results-annotation-projection-llm}, excluding those that underperform or exhibit stability issues. Notably,
only two models use the correct set of 14 tags in a consistent manner, and only in the HTML task: Gemma3 (27.4B parameters) and Mistral (122.6B). Surprisingly, the smaller Gemma3 outperforms Mistral achieving a $0.57$ and $0.58$ \fone on French$\rightarrow$Italian and Italian$\rightarrow$French projection. However, Gemma3 fails the second task: it does not complete Italian$\rightarrow$French due to stability issues and performs worse than AweSome-a. In the IOB task, only Llama3 (70.6B) uses the correct 14 tags for French$\rightarrow$Italian reaching $0.67$ \fone, which is only marginally better than AweSome-a on the same language pair. If we allow one additional and wrong tag for Italian$\rightarrow$French, then LLama3 (70.6B) remains the best performing with a $0.62$ \fone, though it underperforms AweSome-a on the same language pair. These results are somewhat surprising. Not only do they suggest that none of the tested models, with the possible exception of Gemma3, are pretrained on similar tasks, but they also indicate that even very large models struggle to follow simple but unconventional (probably out-of-distribution) instructions, such as ``do not introduce new tags''.  

% ----------------------------------------------------------------
\section*{Conclusions and Future Work}\label{sec:conclusion}
We introduce a novel French--Italian text benchmark for the development and 
evaluation of information extraction methods in the wastewater management domain. 
We provide a baseline for future developments using state-of-the-art, LLM-based NER methods, and explore various annotation projection techniques, including approaches based on LLMs with over 120B parameters. 
The experimental results clearly indicate promising directions for future research. First, we plan to extend the \starwars corpus collecting additional documents, particularly important for Italian, currently lacking robust pre-trained LLMs. Second, we aim to further explore the generalization capabilities of LLMs in automatic annotation. Third, in view of extending the corpus to other languages such as English, we will continue investigating techniques for annotation projection. Lastly, despite their low performance, we will continue to investigate the use of decoder-only LLMs for annotation projection, as they may represent a valid solution for low-resources scenarios such as the domain we have studied in this work.

% ----------------------------------------------------------------
\section*{Acknowledgment}
This research has been partially supported by the EU project \starwars -- STormwAteR and WastewAteR networkS heterogeneous data AI-driven management (GA 101086252) and by the ANR project CROQUIS (GA ANR-21-CE23-0004). The experiments were carried out using resources made available by the ISTI-Cloud infrastructure (CNR-ISTI, Pisa, Italy), the HPC4AI center (University of Turin,  Italy), and the Grid'5000 testbed (Inria,  France).

 ----------------------------------------------------------------
%\bibliographystyle{unsrt}
%\bibliography{bibliography}

\begin{thebibliography}{10}

\bibitem{10.1093/database/baw068}
Jiao Li et~al.
\newblock {BioCreative V CDR task corpus: a resource for chemical disease
  relation extraction}.
\newblock {\em Database}, 2016, 2016.

\bibitem{luan-etal-2018-multi}
Yi~Luan, Luheng He, Mari Ostendorf, and Hannaneh Hajishirzi.
\newblock Multi-task identification of entities, relations, and coreference for
  scientific knowledge graph construction.
\newblock In {\em Proc. of the 2018 Conf. on Empirical Methods in Natural
  Language Processing}, pages 3219--3232, 2018.

\bibitem{10.3897/BDJ.10.e89481}
Nora Abdelmageed et~al.
\newblock Biodiv\uppercase{NERE}: Gold standard corpora for named entity
  recognition and relation extraction in the biodiversity domain.
\newblock {\em Biodiversity Data Journal}, 10, 2022.

\bibitem{btz682}
Jinhyuk Lee et~al.
\newblock Bio\uppercase{BERT}: a pre-trained biomedical language representation
  model for biomedical text mining.
\newblock {\em Bioinformatics}, 36(4):1234--1240, 2019.

\bibitem{8965108}
Xin Yu, Wenshen Hu, Sha Lu, Xiaoyan Sun, and Zhenming Yuan.
\newblock {Bio\uppercase{BERT} Based Named Entity Recognition in Electronic
  Medical Record}.
\newblock In {\em 10th Int'l. Conf. on Information Technology in Medicine and
  Education (ITME)}, pages 49--52, 2019.

\bibitem{KOSPRDIC2024102970}
Miloš Košprdić, Nikola Prodanović, Adela Ljajić, Bojana Bašaragin, and
  Nikola Milošević.
\newblock From zero to hero: Harnessing transformers for biomedical named
  entity recognition in zero- and few-shot contexts.
\newblock {\em Artificial Intelligence in Medicine}, 156, 2024.

\bibitem{w16213075}
Yi~Ren, Tianyi Zhang, Xurong Dong, Weibin Li, Zhiyang Wang, Jie He, Hanzhi
  Zhang, and Licheng Jiao.
\newblock Watergpt: Training a large language model to become a hydrology
  expert.
\newblock {\em Water}, 16 (21), 2024.

\bibitem{malmasi2022multiconerlargescalemultilingualdataset}
Shervin Malmasi, Anjie Fang, Besnik Fetahu, Sudipta Kar, and Oleg Rokhlenko.
\newblock {MultiCoNER: A Large-scale Multilingual dataset for Complex Named
  Entity Recognition}.
\newblock {\em arXiv}, 2022.

\bibitem{zhu2024multilinguallargelanguagemodels}
Shaolin Zhu, Supryadi, Shaoyang Xu, Haoran Sun, Leiyu Pan, Menglong Cui,
  Jiangcun Du, Renren Jin, António Branco, and Deyi Xiong.
\newblock Multilingual large language models: A systematic survey.
\newblock {\em arXiv}, 2024.

\bibitem{cherfi_weir-p_2021}
Nanée Chahinian, Thierry Bonnabaud La~Bruyère, Francesca Frontini, Carole
  Delenne, Julien Marin, Rachel Panckhurst, Mathieu Roche, Lucile Sautot,
  Laurent Deruelle, and Maguelonne Teisseire.
\newblock {WEIR}-{P}: {An} {Information} {Extraction} {Pipeline} for the
  {Wastewater} {Domain}.
\newblock In {\em Proc. of Research Challenges in Information Science, RCIS},
  pages 171--188, 2021.

\bibitem{H0VXH0_2020}
Nanée Chahinian et~al.
\newblock {Gold Standard du projet MeDo}, 2020.

\bibitem{geographique_geostandard_2019}
Géostandard {Réseaux} d’adduction d’eau potable et d’assainissement
  ({RAEPA}) v1.2, 2019.

\bibitem{haydar_ontology_2024}
Batoul Haydar, Claude Pasquier, Umberto Straccia, and Nanée Chahinian.
\newblock An ontology based data access framework for sewer network data.
\newblock {\em Submitted to Automation in Construction}, 2024.

\bibitem{TjongKimSangBuchholz2000Introduction}
Erik~F. Tjong Kim~Sang and Sabine Buchholz.
\newblock Introduction to the {{CoNLL-2000 Shared Task Chunking}}.
\newblock In {\em Fourth {{Conf.}} on {{Computational Natural Language
  Learning}} and the {{Second Learning Language}} in {{Logic Workshop}}}, 2000.

\bibitem{devlin2019bert}
Jacob Devlin et~al.
\newblock Bert: Pre-training of deep bidirectional transformers for language
  understanding.
\newblock In {\em Proc. of the Conf. of the North American chapter of the
  \uppercase{ACL}: human language technologies, Vol. 1}, pages 4171--4186,
  2019.

\bibitem{Martin2020}
Louis Martin et~al.
\newblock Camembert: a tasty french language model.
\newblock In {\em Proc. of the 58th Annual Meeting of the \uppercase{ACL}}.
  \uppercase{ACL}, 2020.

\bibitem{garcia2022t}
Iker Garc{\'\i}a-Ferrero, Rodrigo Agerri, and German Rigau.
\newblock T-projection: High quality annotation projection for sequence
  labeling tasks.
\newblock {\em arXiv preprint arXiv:2212.10548}, 2022.

\bibitem{sabet2020simalign}
Masoud~Jalili Sabet et~al.
\newblock Simalign: High quality word alignments without parallel training data
  using static and contextualized embeddings.
\newblock In {\em Findings of the \uppercase{ACL}: EMNLP 2020}, 2020.

\bibitem{DouNeubig2021Worda}
Zi-Yi Dou and Graham Neubig.
\newblock Word {{Alignment}} by {{Fine-tuning Embeddings}} on {{Parallel
  Corpora}}.
\newblock In {\em Proc. of the 16th {{Conf.}} of the {{European Chapter}} of
  the \uppercase{ACL}: {{Main Volume}}}, 2021.

\bibitem{bojanowski2017enriching}
Piotr Bojanowski, Edouard Grave, Armand Joulin, and Tomas Mikolov.
\newblock Enriching word vectors with subword information.
\newblock {\em Trans. of the \uppercase{ACL}}, 5, 2017.

\bibitem{conneau2020unsupervised}
Alexis Conneau et~al.
\newblock Unsupervised cross-lingual representation learning at scale.
\newblock In {\em Proc. of the 58th Annual Meeting of the \uppercase{ACL}},
  2020.

\bibitem{touvron2023llama}
Hugo Touvron et~al.
\newblock Llama 2: Open foundation and fine-tuned chat models.
\newblock {\em arXiv preprint arXiv:2307.09288}, 2023.

\bibitem{grattafiori2024llama}
Aaron Grattafiori et~al.
\newblock The llama 3 herd of models.
\newblock {\em arXiv preprint arXiv:2407.21783}, 2024.

\bibitem{abdin2024phi}
Marah Abdin et~al.
\newblock Phi-4 technical report.
\newblock {\em arXiv preprint arXiv:2412.08905}, 2024.

\bibitem{seqeval}
Hiroki Nakayama.
\newblock {seqeval}: A python framework for sequence labeling evaluation, 2018.

\bibitem{semeval13}
Isabel Segura-Bedmar et~al.
\newblock {S}em{E}val-2013 task 9 : Extraction of drug-drug interactions from
  biomedical texts.
\newblock In {\em Proc. of the 7th International Workshop on Semantic
  Evaluation}, pages 341--350, 2013.

\end{thebibliography}

\end{document}